\documentclass{article}
\usepackage{graphicx}
\usepackage{biblatex}
\usepackage[preprint, nonatbib]{tackling_climate_workshop_style}

\usepackage[utf8]{inputenc} 
\usepackage[T1]{fontenc}    
\usepackage{hyperref}       
\usepackage{url}            
\usepackage{booktabs}       
\usepackage{amsfonts}       
\usepackage{nicefrac}       
\usepackage{microtype}      

\title{Trend and Thoughts: Understanding Climate Change Concern using Machine Learning and Social Media Data}

\author{%
  Zhongkai Shangguan \thanks{The two authors contribute equally.}\\
  Department of Electrical and Computer Engineering\\
  Boston University\\
  Boston, MA 02215 \\
  \texttt{sgzk@bu.edu} \\
   \And
   Zihe Zheng \textsuperscript{\rm *}\\
   Goergen Institute for Data Science \\
   University of Rochester \\
   Rochester, NY 14627 \\
   \texttt{zzheng18@u.rochester.edu} \\
   \AND
   Lei Lin \\
   Goergen Institute for Data Science \\
   University of Rochester \\
   Rochester, NY 14627 \\
   \texttt{Lei.Lin@rochester.edu} \\
}

\begin{document}

\maketitle

\begin{abstract}
Nowadays social media platforms such as Twitter provide a great opportunity to understand public opinion of climate change comparing to traditional survey methods. In this paper, we constructed a massive climate change Twitter dataset and conducted comprehensive analysis using machine learning. By conducting topic modeling and natural language processing, we show the relationship between the number of tweets about climate change and major climate events; the common topics people discuss about climate change; and the trend of sentiment. Our dataset was published on Kaggle (\url{https://www.kaggle.com/leonshangguan/climate-change-tweets-ids-until-aug-2021}) and can be used in further research.
\end{abstract}

\section{Introduction and Related Work}

In recent years, the issue of climate change has raised widespread concern. At the same time, with the rapid development of the internet and communication technology, more and more people tend to express their opinions through social media, which makes it a great resource to study public opinion of climate change. As millions of people are releasing massive amounts of data every day on social media, obtaining and analyzing these data become extremely challenging. Fortunately, the development of computing hardware and machine learning algorithms have enabled us to analyze and make use of these social media data. In 2019, Rolnick et al. \cite{rolnick2019tackling} discussed how machine learning can potentially be a powerful tool in tackling climate change issues. In particular, they pointed out how machine learning could extract key information from big data, which can be used in disaster response, as well as smart city construction. 
An et al. \cite{an2014tracking} utilized a bag-of-words technique on Twitter data and performed sentiment analysis, which reveals that negative sentiment is mainly related to major climate events. Fownes et al. \cite{fownes2018twitter} gave a comprehensive review on how climate change was discussed on Twitter and suggested future areas of study. In 2019, Littman et al. published the Climate Change Tweets Ids Dataset \cite{DVN/5QCCUU_2019} which included millions of tweets related to climate change. More specific topics such as Super Cyclonic Storm Amphan and Drought Impacts on Twitter were also studied \cite{crayton2020narratives, zhangtweetdrought}.

In this study, we propose 12 keywords to collect discussion threads related to climate change on Twitter. We analyze the trend, conduct topic modeling and sentiment analysis using ML models for Natural Language Processing (NLP). Our contributions can be summarized as follows: First, we build an open-sourced climate change Twitter dataset. Second, by performing big data analysis, we show the trend, common topics, and people's sentiments about climate change on Twitter.

\section{Data Collection and Preprocessing}

We extracted tweets from April 01, 2019 to August 31, 2021 by snscrape \cite{snscrape}, using 12 keywords as the Harvard Dataverse \cite{DVN/5QCCUU_2019}, i.e., \textit{\#climatechange, \#climatechangeisreal, \#actonclimate, \#globalwarming, \#climatechangehoax, \#climatedeniers, \#climatechangeisfalse, \#globalwarminghoax, \#climatechangenotreal, climate change, global warming, climate hoax}. Four tweet variable were kept including username, published date, tweet ID and tweet content. After removing the duplicated tweets, a new climate change Twitter dataset containing 15,075,535 tweets is generated. 

Note that there is another climate change dataset from Harvard Dataverse \cite{DVN/5QCCUU_2019} and retrieved using Hydrator \cite{Hydrator}, which contains extra 6,430,632 tweets \footnote{The dataset retrieved is smaller than the original one since some tweets/users might be deleted.} from September 21, 2017 to May 17, 2019, with two gaps from October to November, 2018, and from January to April, 2019. To the best of our knowledge, our dataset is the largest and most recent climate change Twitter dataset. 

\section{Method}

\subsection{Topic Modeling}

We apply topic modeling using the Latent Dirichlet Allocation (LDA) model to summarize the topics that are discussed about climate change. LDA is a generative probabilistic model that can be used to find topics that a document belongs to based on the words in it \cite{blei2003latent}. The algorithm first calculates the probabilities of various words appearing in each topic, while the words with the highest possibility of appearing in that topic defines the topic. Ten percent of the tweets was randomly chosen from the new climate change Twitter dataset, which contains 1,507,554 tweets. We use the spaCy software \cite{spacy} to lemmatize the words, an all sentences are transformed to lowercase letters. To prevent the influence of stop words, we removed the stop words (e.g. the, but, we) using the Natural Language Toolkit. Since most of the tweets are about climate change, we decided to also remove the keyword "climate change" from the dataset, so that the result of the topic modeling includes more interesting and diverse topics.

\subsection{Sentiment Analysis}

Sentiment analysis is a widely used technology in social media analysis. It is able to automatically determine whether the author is in favor of (positive), against (negative), or neutral towards a topic or viewpoint \cite{aldayel2021stance,  mohammad2016semeval}. 
In recent years, with the introduction of machine learning and deep neural networks into NLP, a large number of NLP models designed for different situations have emerged. RoBERTa, in full Robustly Optimized BERT Pre-training Approach \cite{liu2019roberta} , is an advanced NLP model which incorporated improved training task and data generation methods to achieve the state of the art. In our research, we use a pre-trained RoBERTa-base model which was re-trained on the Semeval2017 dataset\cite{rosenthal2017semeval} and reported a 72.6 macro-averaged recall \cite{barbieri2020tweeteval}. The sentiments of contents are classified with three labels: negative, neutral and positive.

\section{Results and Discussion}

\subsection{Trend}

\begin{figure}
  \centering
  \includegraphics[width=\linewidth]{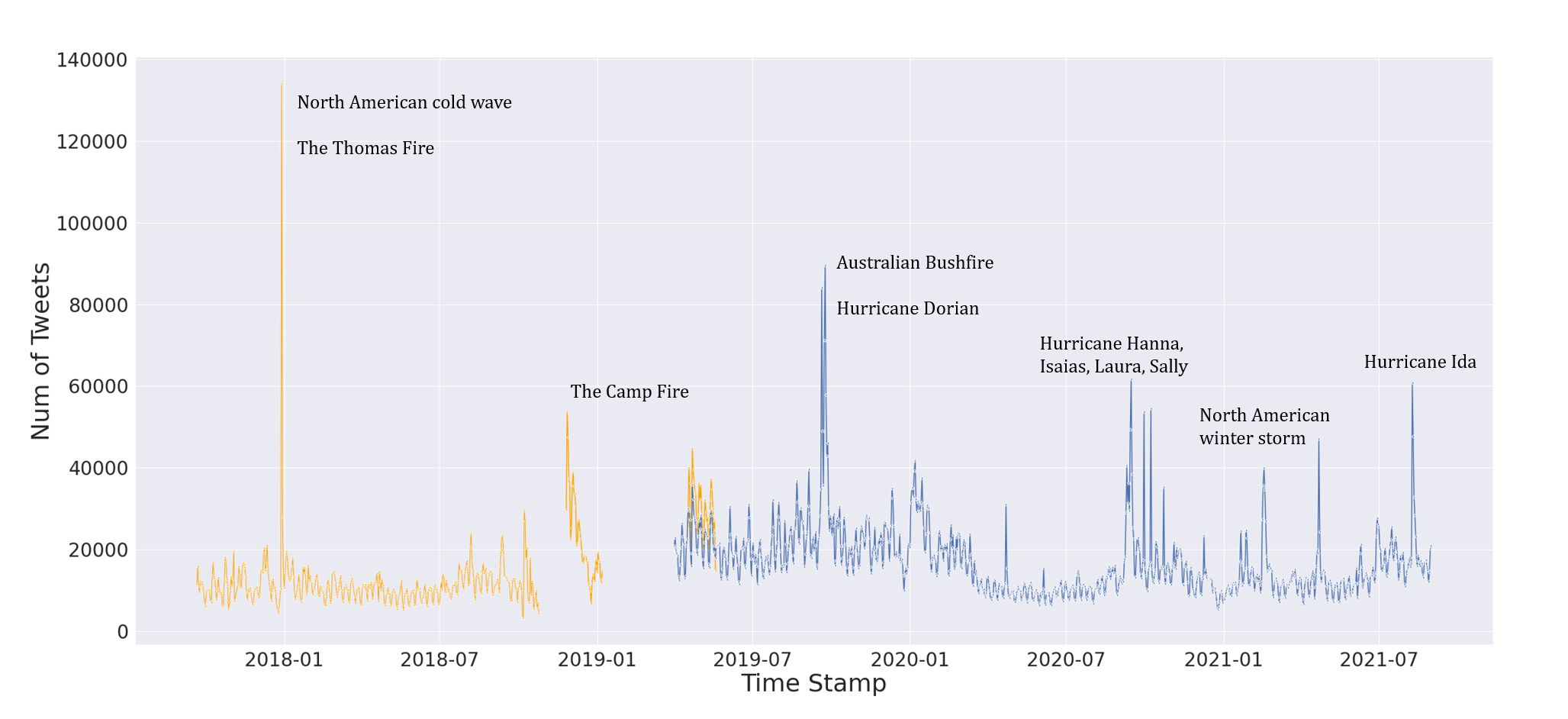}
  \caption{\label{fig:trend} Trend of the discussion of climate change, annotated with major climate events}
\end{figure}

We show the number of tweets that are related to climate change sent by different people in Fig. \ref{fig:trend}, where the yellow line represents the Harvard Dataverse \cite{DVN/5QCCUU_2019} and the blue line represents our new dataset. The two lines overlap in April 2019, where data is collect in both datasets. At each spike of discussion, we look up major climate events during that time, and annotate the spike by the events. The number of tweets about climate change is dependent on climate events as most of the spikes are accompanied by one or several events, except the high points in April 2020, and April 2021, where the Earth Day is celebrated on social media.

\begin{figure}
  \centering
  \includegraphics[width=\linewidth]{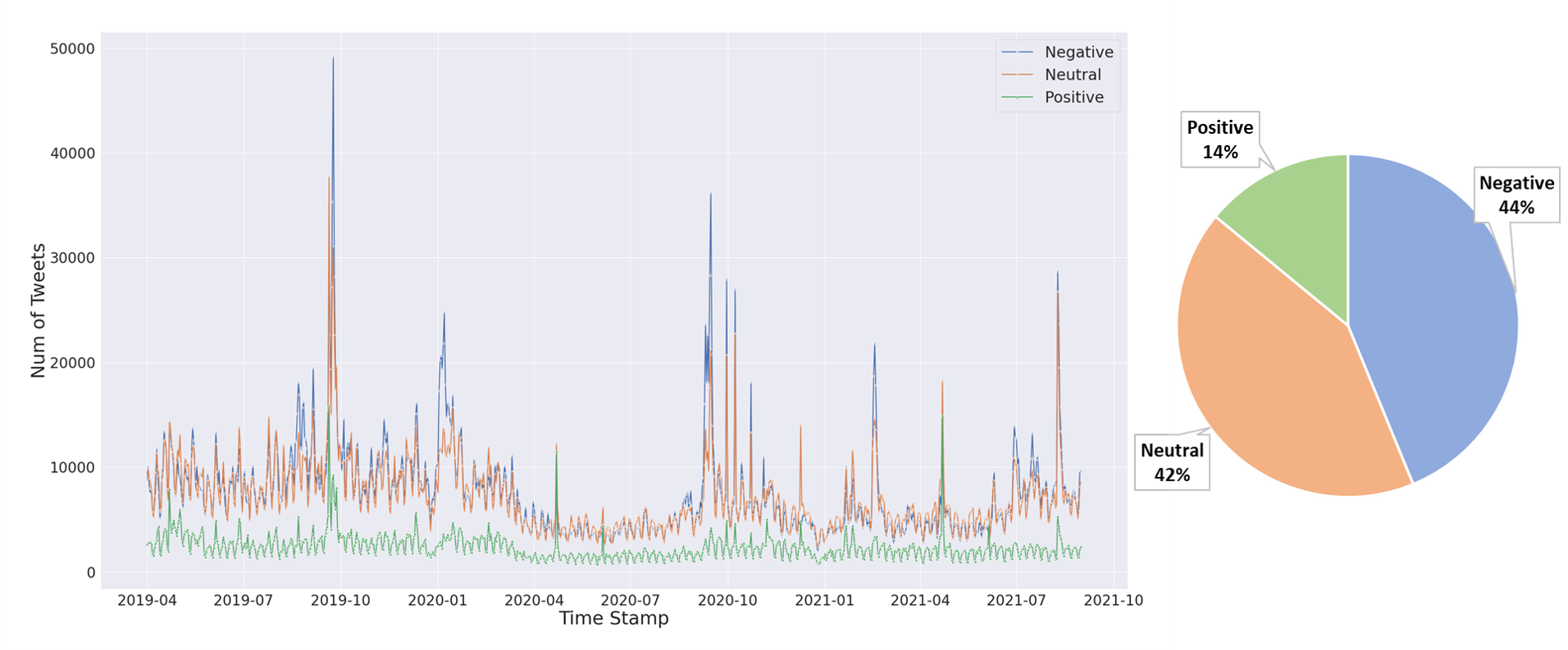}
  \caption{\label{fig:sentiment} Left: People's sentiment across time; Right: Proportion of different sentiments}
\end{figure}

\subsection{Topic Modeling}

Topic modeling using LDA was performed on ten percent of the new dataset to identify common topics discussed on Twitter about climate change. The number of topic was set to 15 to achieve the least overlap between topics while all interesting topics are represented. We show the 15 topics about climate change on Twitter, and the top 10 keywords of each topic in Table \ref{tab: topics}. 

Many tweets are found to be discussing the importance of climate change. People tweet about the love of our children (topic 5), say that we should save lives (topic 6), persuade climate change deniers (topic 8), show scientific results (topic 10), call on people to fight and protest against climate change (topic 11 and topic 14), and comment about information from media (topic 12). Other discussions focus on certain aspects of climate change, such as extreme weather (topic 9), wildfire (topic 13), and global warming (topic 15). There are also voices about how we can solve the problem of climate change: reducing carbon emission (topic 2), practicing sustainable development (topic 3), and protecting nature from pollution (topic 4).
Since our study period spans from April 2019 to August 2021, COVID-19 is found to be another major topics of the climate change discussion (topic 7). The pandemic has posed a public health challenge to the world, resulting in economic downturns in many countries. We can infer that the pandemic has also influenced people's concern of climate change. Topic 1 is found to be entirely about the presidential election between Trump and Biden, indicating that climate change policy is an issue that is very important to the number of votes in the general election. 

\begin{table}[]
\scalebox{0.82}{
\begin{tabular}{|c|l|c|}
\hline
Topic & \multicolumn{1}{c|}{Description}                                                                                                                                         & Keywords                                                                                                                          \\ \hline
1     & \begin{tabular}[c]{@{}l@{}}Discussion about the presidential election between \\ Trump and Biden.\end{tabular}                                                           & \begin{tabular}[c]{@{}c@{}}Trump, plan, policy, vote, support, \\ Biden, deal, American, president, party\end{tabular}            \\ \hline
2     & \begin{tabular}[c]{@{}l@{}}Company should reduce carbon emission and \\ energy usage.\end{tabular}                                                                       & \begin{tabular}[c]{@{}c@{}}Energy, emission, carbon, power, reduce, \\ oil, company, green, industry, China\end{tabular}          \\ \hline
3     & \begin{tabular}[c]{@{}l@{}}We need to work together as a community to figure \\ out a sustainable solution.\end{tabular}                                                 & \begin{tabular}[c]{@{}c@{}}Community, project, solution, learn, sustainability, \\ city, tackle, discuss, work, role\end{tabular} \\ \hline
4     & \begin{tabular}[c]{@{}l@{}}We should protect our environment, water, food, \\ and nature from pollution.\end{tabular}                                                    & \begin{tabular}[c]{@{}c@{}}Environment, water, pollution, tree, food, \\ forest, plant, protect, nature, air\end{tabular}         \\ \hline
5     & \begin{tabular}[c]{@{}l@{}}People love and care about their kid and hope to \\ leave them with a better climate.\end{tabular}                                            & \begin{tabular}[c]{@{}c@{}}People, child, care, kid, feel, \\ love, women, give, leave, thing\end{tabular}                        \\ \hline
6     & \begin{tabular}[c]{@{}l@{}}People in the world are all facing the problem of \\ climate change and we should save the planet and the \\ lives living on it.\end{tabular} & \begin{tabular}[c]{@{}c@{}}World, people, planet, problem, country, \\ end, save, money, life, live\end{tabular}                  \\ \hline
7     & \begin{tabular}[c]{@{}l@{}}The government should act on the health crisis brought \\ by COVID-19.\end{tabular}                                                           & \begin{tabular}[c]{@{}c@{}}Action, covid, future, issue, address, \\ crisis, government, health, threat, face\end{tabular}        \\ \hline
8     & \begin{tabular}[c]{@{}l@{}}Debate between people who believe in science and\\ climate change deniers.\end{tabular}                                                       & \begin{tabular}[c]{@{}c@{}}Make, science, scientist, man, hoax, \\ real, fact, deny, denier, lie\end{tabular}                     \\ \hline
9     & \begin{tabular}[c]{@{}l@{}}The rising temperature has caused extreme weathers \\ like floods.\end{tabular}                                                               & \begin{tabular}[c]{@{}c@{}}Weather, due, increase, high, rise, \\ temperature, flood, level, extreme, record\end{tabular}         \\ \hline
10    & \begin{tabular}[c]{@{}l@{}}Discussions of results shown by research papers and \\ reports.\end{tabular}                                                                  & \begin{tabular}[c]{@{}c@{}}Impact, effect, report, read, human, \\ show, find, risk, study, research\end{tabular}                 \\ \hline
11    & Fighting hard to combat climate change.                                                                                                                                  & \begin{tabular}[c]{@{}c@{}}Fight, work, good, put, move, \\ home, big, bring, give, hard\end{tabular}                             \\ \hline
12    & Discussions of media coverage about climate change.                                                                                                                      & \begin{tabular}[c]{@{}c@{}}Talk, news, point, good, issue, \\ question, debate, lot, medium, hear\end{tabular}                    \\ \hline
13    & \begin{tabular}[c]{@{}l@{}}Discussions about wildfire in Australia and whom to \\ blame for it.\end{tabular}                                                             & \begin{tabular}[c]{@{}c@{}}Year, time, fire, bad, start, \\ happen, long, Australia, state, blame\end{tabular}                    \\ \hline
14    & Calling on people to join protests about climate change.                                                                                                                 & \begin{tabular}[c]{@{}c@{}}Day, today, week, watch, join, \\ action, protest, activist, check, story\end{tabular}                 \\ \hline
15    & Global warming is real and we should try to stop it.                                                                                                                     & \begin{tabular}[c]{@{}c@{}}Global, warming, earth, real, stop, \\ warm, call, change, cool, sun\end{tabular}                      \\ \hline
\end{tabular}
}

\caption{\label{tab: topics} Topics about climate change on Twitter}
\end{table}

\subsection{Sentiment Analysis}

We demonstrate the trend and proportion of people's sentiment towards climate change in Fig. \ref{fig:sentiment}. As it is shown, many of tweets (44\%) have shown negative attitudes towards climate change, while nearly the same amount of tweets are labeled as neutral. However, only 14\% tweets show positive sentiment, which is about one third as many compared with the negative/neutral tweets. 
Over time, the negative and neutral sentiment show similar trends, while only at the time of major climate events that the number of negative sentiment tweets rises more rapidly and largely exceeds neutral tweets.
The positive sentiment only dominants the discussion on April 22, 2020 and 2021, when the $50^{th}$ and $51^{st}$ anniversaries of Earth Day were celebrated online in the form of digital media.

\section{Conclusion}
Through temporal analysis of the discussions of climate change on Twitter, we find a dependency that the number of tweets about climate change usually rise on major climate events. Topic modeling on the new climate change dataset show that people discuss about the importance of climate change, major climate events, and how to tackle the problem of climate change. We also find political and COVID-19 related topics in the discussion of climate change. Further sentiments analyses over time reveal that tweets with a negative sentiment typically rise higher at the time of these events than neutral and positive ones. Overall, the negative and neutral tweets compose 44 and 42 percent of total tweets, while positive tweets are much less. This study sheds light on the content and trend of public opinions and discussions about climate change, and the results of the temporal, topic, and sentiment analysis could be used as a reference to future campaign and to understand people’s views on climate change. We published our dataset on Kaggle \cite{climate_2021} which makes it convenient for researchers to conduct further research.

\printbibliography

\end{document}